# Prediction with Advice of Unknown Number of Experts

Alexey Chernov and Vladimir Vovk

Computer Learning Research Centre, Department of Computer Science
Royal Holloway, University of London, Egham, Surrey TW20 0EX, UK
{chernov,vovk}@cs.rhul.ac.uk

October 9, 2018

## Abstract

In the framework of prediction with expert advice, we consider a recently introduced kind of regret bounds: the bounds that depend on the effective instead of nominal number of experts. In contrast to the Normal-Hedge bound, which mainly depends on the effective number of experts and also weakly depends on the nominal one, we obtain a bound that does not contain the nominal number of experts at all. We use the defensive forecasting method and introduce an application of defensive forecasting to multivalued supermartingales.

## 1 Introduction

We consider the problem of prediction with expert advice (PEA) and its variant, decision-theoretic online learning (DTOL). In the PEA framework (see [3] for details, references and historical notes), at each step Learner gets decisions (also called predictions) of several Experts and must make his own decision. Then the environment generates an outcome and a (real-valued) loss is calculated for each decision as a known function of decision and outcome. The difference between cumulative losses of Learner and one of Experts is the regret to this Expert. Learner aims at minimizing his regret to Experts, for any sequence of Expert decisions and outcomes.

In DTOL, introduced in [8], Learner's decision is a probability distribution on a finite set of actions. Then each action incurs a loss (the vector of the losses can be regarded as the outcome), and Learner suffers the loss equal to the expected loss over all actions (according to the probabilities from his decision). The regret is the difference between the cumulative losses of Learner and one of the actions. One can interpret each action as a rigid Expert that always suggests this action. A precise connection between the DTOL and PEA frameworks will be described in Section 2.

Usually Learner is required to have small regret to all Experts. In other words, a strategy for Learner must have a guaranteed upper bound on Learner's regret to the best Expert (one with the minimal loss). In this paper we deal with another kind of bound, recently introduced in [4]. It captures the following intuition. Generally speaking, the more Experts (or actions, in the DTOL terminology) Learner must take into account, the worse his performance will be. However, assume that each Expert has several different names, so Learner is given a lot of identical advice. It seems natural that the loss of Learner is big if there is a real controversy between Experts (or a real difference between actions), and small if most of the Experts agree with each other. So a competent regret bound should depend on the real number of Experts instead of the nominal one. Another example: assume that all the actions are different, but many of them are good — there are many ways to achieve some goal. Then Learner has less space to make a mistake and to select a bad action. Again it seems that a competent regret bound should depend on the fraction of the good actions rather than the nominal number of actions.

If the effective number of actions (Experts) is significantly less than the nominal one, one can loosely say that the number of actions is unknown in this setting. The following regret bound obtained in [4] for their NormalHedge algorithm holds for this case:

$$L_T \leq L_T^\epsilon + O\left(\sqrt{T \ln \frac{1}{\epsilon}} + \ln^2 N\right),\qquad(1)$$

where $N$ is the nominal number of actions, $L_T$ is the cumulative loss of Learner after step $T$ and $L_T^\epsilon$ is the value such that at least $\epsilon$-fraction of actions have smaller or equal cumulative loss after step $T$ (or $L_T^\epsilon$ can be interpreted as the loss of $\epsilon N$-th best action). It is important that the bound holds uniformly for all $\epsilon$ and $T$ and the algorithm does not need to know them in advance. The number $\frac{1}{\epsilon}$ plays the role of the *effective* number of actions. The bound shows, in a sense, that the NormalHedge algorithm can work even if the number of actions is not known.

Our main result (Theorem 9) is the following bound for a new algorithm:

$$L_T \leq L_T^\epsilon + 2\sqrt{T \ln \frac{1}{\epsilon}} + 7\sqrt{T}.$$

This bound is also uniform in $T$ and $\epsilon$. In contrast to (1), our bound does not depend on the nominal number of actions, whereas (1) contains a term $O(\ln^2 N)$. So it is the first (as far as we know) bound strictly in terms of the effective number of actions. Our bound has a simpler structure, but it is generally incomparable to the (precise) bound for Normal Hedge from [4] (see Subsection 4.2 for discussion of different known bounds). Also our bound can be easily adapted to internal regret (see [12] for definition). We describe the application to internal regret in Subsection 4.3.

Our bound is obtained with the help of the defensive forecasting method (DF). The DF is based on bounding the growth of some supermartingale (a kind of potential function). In [5], the DF was used to obtain bounds of the form $L_T \leq cL_T^n + a$, where $c$ and $a$ are some constants. For our form of bounds, we need a new variation of the DF and a new sort of supermartingales. So we introduce the notion of multivalued supermartingale and prove a boundedness result for them (Lemmas 2 and 3). (This result is of certain independent interest: for example, it helps to get rid of additional Assumption 3 in Theorem 3 in [5].)

The paper is organized as follows. In Section 2 we describe the setup of prediction with expert advice and of decision-theoretic framework online learn-ing, and define the $\epsilon$-quantile regret. In Section 3 we describe the Defensive Forecasting Algorithm, define multivalued supermartingales and discuss their properties, and introduce supermartingales of a specific form that are based on Hoeffding inequality. In Subsection 4.1 we prove two loss bounds on the $\epsilon$-quantile regret, and in Subsection 4.2 we compare them with the bound for the NormalHedge algorithm and with other known bounds. In Subsection 4.3 we show how these bounds can be transformed into bounds on the internal regret. In the last subsection we describe a toy example of an algorithm that guarantees bounds for two very different loss functions simultaneously.

## 2 Notation and Setup

Vectors with coordinates $p_1, \ldots, p_N$ are denoted by an arrow over the letter: $\vec{p} = (p_1, \ldots, p_N)$. For any natural $N$, by $\Delta_N$ we denote the standard simplex in $\mathbb{R}^N$: $\Delta_N = \{\vec{p} \in [0,1]^N \mid \sum_{n=1}^N p_n = 1\}$. By $\vec{p} \cdot \vec{q}$ we denote the scalar product: $\vec{p} \cdot \vec{q} = \sum_{n=1}^N p_n q_n$.

**Protocol 1** Decision-theoretic framework for learning

$L_0 := 0$.
$L_0^n := 0$, $n = 1, \ldots, N$.
**for** $t = 1, 2, \ldots$ **do**
  Learner announces $\vec{\gamma}_t \in \Delta_N$.
  Reality announces $\vec{\omega}_t \in [0,1]^N$.
  $L_t := L_{t-1} + \vec{\gamma}_t \cdot \vec{\omega}_t$.
  $L_t^n := L_{t-1}^n + \omega_{t,n}$, $n = 1, \ldots, N$.
**end for**

The decision-theoretic framework for online learning (DTOL) was introduced in [8]. DTOL protocol is given as Protocol 1. The Learner has $N$ available actions, and at each step $t$ he must assign probability weights $\gamma_{t,1}, \ldots, \gamma_{t,N}$ to these actions. Then each action suffers a loss $\omega_{t,n}$, and Learner's loss is the expected loss over all actions according to the weights he assigned. Learner's goal is to keep small his regret $R_t^n = L_t - L_t^n$ to any action $n$, independent of the losses.

**Protocol 2** Prediction with Expert Advice

$L_0 := 0$.
$L_0^n := 0$, $n = 1, \ldots, N$.
**for** $t = 1, 2, \ldots$ **do**
  Expert $n$ announces $\gamma_t^n \in \Gamma$, $n = 1, \ldots, N$.
  Learner announces $\gamma_t \in \Gamma$.
  Reality announces $\omega_t \in \Omega$.
  $L_t := L_{t-1} + \lambda(\gamma_t, \omega_t)$.
  $L_t^n := L_{t-1}^n + \lambda(\gamma_t^n, \omega_t)$, $n = 1, \ldots, N$.
**end for**

DTOL can be regarded as a special case of prediction with expert advice (PEA), as explained below. The PEA protocol is given as Protocol 2. The game is specified by the set of outcomes $\Omega$, the set of decisions $\Gamma$ and the loss function $\lambda : \Gamma \times \Omega \to \mathbb{R}$. The game is played repeatedly by Learner having access to decisions made by a pool of Experts. At each step, Learner is given $N$ Experts' decisions and is required to come out with his own decision. The loss $\lambda(\gamma, \omega)$ measures the discrepancy between the decision $\gamma$ and the outcome $\omega$. $L_t$ is Learner's cumulative loss over the first $t$ steps, and $L_t^n$ is the $n$-th Expert's cumulative loss over the first $t$ steps. The goal of Learner is the same: to keep small his regret $R_t^n = L_t - L_t^n$ to any Expert $n$, independent of Experts' moves and the outcomes.

As defined in [4] (for DTOL), the *regret to the top $\epsilon$-quantile* (at step $T$) is the value $R_T^\epsilon$ such that there are at least $\epsilon N$ actions (the fraction at least $\epsilon$ of all Experts) with $R_T^n \geq R_T^\epsilon$. Or, equivalently, $R_T^\epsilon = L_T - L_T^\epsilon$ where $L_T^\epsilon$ is a value such that at least $\epsilon N$ actions (the fraction at least $\epsilon$ of all Experts) has the loss $L_T^n$ less than $L_T^\epsilon$.

A uniform bound on $R_T^\epsilon$ (in other words, a bound on Learner's loss $L_T$ in terms of $L_T^\epsilon$) that holds for all $\epsilon$ is more general than the standard best Expert bounds. The latter can be obtained as a special case for $\epsilon = 1/N$. For this reason, it is natural to call the value $1/\epsilon$ the *effective* number of actions: a bound on $R_T^\epsilon$ can be considered as the best Expert bound in an imaginary game against $1/\epsilon$ Experts.

Let us say what games $(\Omega, \Gamma, \Lambda)$ we consider in this paper. For any game $(\Omega, \Gamma, \lambda)$, we call $\Lambda = \{g \in \mathbb{R}^\Omega \mid \exists \gamma \in \Gamma \, \forall \omega \in \Omega \, g(\omega) = \lambda(\gamma, \omega)\}$ the *prediction set*. The prediction set captures most of the information about the game. The prediction set is assumed to be non-empty. In this paper, we consider *bounded convex compact games* only. This means that we assume that the set $\Lambda$ is bounded and compact, and the superprediction set $\Lambda + [0, \infty]^\Omega$ is convex, that is, for any $g_1, \ldots, g_K \in \Lambda$ and for any $p_1, \ldots, p_K \in [0,1]^K$, $\sum_{k=1}^K p_k = 1$, there exists $g \in \Lambda$ such that $g(\omega) \leq \sum_{k=1}^K p_k g_k(\omega)$ for all $\omega \in \Omega$. For such games, we assume without loss of generality that $\Lambda \subseteq [0,1]^\Omega$ (we always can scale the loss function).

For DTOL as a special case of PEA, the outcome space is $\Omega = [0,1]^N$, the decision space is $\Gamma = \Delta_N$, and the loss function is $\lambda(\vec{\gamma}, \vec{\omega}) = \vec{\gamma} \cdot \vec{\omega}$. Experts play fixed strategies always choosing $\vec{\gamma}_t^n$ such that $\gamma_{t,n}^n = 1$ and $\gamma_{t,k}^n = 0$ for $k \neq n$ (see e.g. [13, Example 7] for more details about this game).

In an important sense the general PEA protocol for the bounded convex games is equivalent to DTOL. Obviously, if some upper bound on regret is achievable in any PEA game then it is achievable in the special case of the DTOL game. To see how to transfer an upper bound from DTOL to a PEA game, let us interpret the decisions $\gamma_t^n$ of Experts and the outcome $\omega_t$ in the PEA game as the outcome $\vec{\omega}_t'$ in DTOL: $\omega_{t,n}' = \lambda(\gamma_t^n, \omega_t)$. If Learner's decision $\gamma_t$ satisfies $\lambda(\gamma_t, \omega_t) \leq \sum_{n=1}^N \gamma_{t,n}' \lambda(\gamma_t^n, \omega_t)$, where $\vec{\gamma}_t'$ is Learner's decision in DTOL, then the regret (at step $t$) in the PEA game will be not greater than the regret in DTOL. It remains to note that, since the game is convex, for any $\vec{\gamma}_t'$ there exists $\gamma_t$ such that $\lambda(\gamma_t, \omega) \leq \sum_{n=1}^N \gamma_{t,n}' \lambda(\gamma_t^n, \omega)$, for any $\omega \in \Omega$.

However, the equivalence between DTOL and PEA is limited. In particular, we can obtain PEA bounds that hold for specific loss functions or classes of loss functions (such as mixable loss functions [13]), and these bounds may be much stronger than the general bounds induced by DTOL.

In this paper, we consider PEA and DTOL in parallel for another reason.

It is sometimes useful to consider a more general variant of Protocol 2 where the number of Experts is infinite (and maybe uncountably infinite): then PEA can be applied to large families of functions as Experts. With the help of our method, we can cope either with DTOL, where the number of actions is finite, or with PEA when $\Omega$ is finite and the number of Experts is arbitrary. So we cannot infer a bound for infinitely many Experts from a DTOL result, but we can obtain a PEA result directly. In the sequel, we will write about $N$ experts, but always allow $N$ to be infinite in the PEA case.

Most of the presentation below is in the terms of PEA but applicable to DTOL as well. We normally hide the difference between PEA and DTOL behind the common notation (DTOL is considered as the game described above). When the difference is important, we give two parallel fragments of a statement or proof.

## 3 Defensive Forecasting and Supermartingales

This section contains the technical results we need to construct our prediction algorithm. They are used in the proofs but not in the theorem statements and discussions in the next section.

### 3.1 Defensive Forecasting

The general structure of the *Defensive Forecasting Algorithm* (DFA) is quite simple. At step $t$, we define a function $f_t : \Gamma \times \Omega \to \mathbb{R}$ (with special properties — see below) and look for $\gamma \in \Gamma$ such that

$$\forall \omega \in \Omega \quad f_t(\gamma, \omega) \leq f_{t-1}(\gamma_{t-1}, \omega_{t-1}), \tag{2}$$

where $f_{t-1}$ is the function defined at the previous step, $\gamma_{t-1}$ is Learner's decision at the previous step, and $\omega_{t-1}$ is the outcome at the previous step. Then $\gamma$ with this property is announced as the next decision of Learner $\gamma_t$.

The choice of $f_t$ may depend on all the previous decisions, outcomes, and on this step Experts' decisions (for PEA), so $f_t = \mathcal{F}(\{\gamma_1^n\}_{n=1}^N, \gamma_1, \omega_1, \ldots, \{\gamma_t^n\}_{n=1}^N)$. Having specified $\mathcal{F}$, we call this strategy of Learner *an application of the DFA to $\mathcal{F}$*.

The algorithm guarantees that the values of $f_t$ do not increase, in particular, after each step the value $f_t(\gamma_t, \omega_t)$ is not greater than some initial value $f_0$. We will choose $\mathcal{F}$ so that the inequality $\mathcal{F}(\{\gamma_1^n\}_{n=1}^N, \gamma_1, \omega_1, \ldots, \{\gamma_t^n\}_{n=1}^N)(\gamma_t, \omega_t) \leq f_0$ implies a loss bound we need.

Also we need to guarantee that the algorithm always can find $\gamma$ satisfying (2). To this end we will choose $\mathcal{F}$ so that the sequence $f_t$ will be a (multivalued) supermartingale as defined in the next subsection.

### 3.2 Multivalued Supermartingales

Let $\Omega$ be a compact metric space. Any finite set $\Omega$ is considered as a metric space with the discrete metric. Let $\mathcal{P}(\Omega)$ be the space of all measures on $\Omega$ supplied with the weak topology.

For any measurable function $g \in \mathbb{R}^\Omega$ and any $\pi \in \mathcal{P}(\Omega)$, denote

$$\mathrm{E}_\pi g = \int_\Omega g(\omega)\pi(d\omega)\,.$$

For finite $\Omega$, this definition reduces to the scalar product:

$$\mathrm{E}_\pi g = \sum_{\omega \in \Omega} g(\omega)\pi(\omega)\,.$$

Let $S$ be an operator that to any sequence $e_1, \pi_1, \omega_1, \ldots, e_{T-1}, \pi_{T-1}, \omega_{T-1}, e_T$, where $\omega_t \in \Omega$, $\pi_t \in \mathcal{P}(\Omega)$, $t = 1, \ldots, T-1$, and $e_t$, $t = 1, \ldots, T$ are some arbitrary values, assigns a function $S_T : \mathcal{P}(\Omega) \to \mathbb{R}^\Omega$. To simplify notation, we will hide the dependence of $S_T$ on all the long argument sequence in the index $T$. We call $S$ a (*game-theoretic*) *supermartingale* if for any sequence of arguments, for any $\pi \in \mathcal{P}(\Omega)$, for $g_{T-1} = S_{T-1}(\pi_{T-1})$ and for $g = S_T(\pi)$ it holds

$$\mathrm{E}_\pi g \leq g_{T-1}(\omega_{T-1})\,. \tag{3}$$

This definition of supermartingale is equivalent to the one given in [5]. We say that supermartingale $S$ is forecast-continuous if every $S_T$ is a continuous function.

The main property of forecast-continuous supermartingales that makes them useful in our context is given by Lemma 1. Originally, a variant of the lemma was obtained by Leonid Levin in 1976. The proof is based on fixed-point considerations, see [10, Theorem 16.1] or [6, Lemma 8] for details.

**Lemma 1.** *Let $\Omega$ be a compact metric space. Let a function $q : \mathcal{P}(\Omega) \times \Omega \to \mathbb{R}$ be continuous as function from $\mathcal{P}(\Omega)$ to $\mathbb{R}^\Omega$. If for all $\pi \in \mathcal{P}(\Omega)$ it holds that*

$$\mathrm{E}_\pi q(\pi, \cdot) \leq C\,,$$

*where $C \in \mathbb{R}$ is some constant, then*

$$\exists \pi \in \mathcal{P}(\Omega)\, \forall \omega \in \Omega \quad q(\pi, \omega) \leq C\,.$$

The lemma guarantees that for any forecast-continuous supermartingale $S$ we can always choose $g_t \in S_t$ such that $g_t(\omega) \leq g_{t-1}(\omega_{t-1})$ for all $\omega$. This is exactly the kind of condition we need for the DFA.

Unfortunately, for the loss bounds we want to obtain, we did not find a suitable forecast-continuous supermartingale. So we define a more general notion of multivalued supermartingale, and prove an appropriate variant of Levin's lemma.

To get the definition of a multivalued supermartingale, we make just three changes in the definition of supermartingale above: operator $S$ depends additionally on $g_t \in S_t(\pi_t)$; $S_T$ is function from $\mathcal{P}(\Omega)$ to non-empty subsets of $\mathbb{R}^\Omega$; the condition (3) holds for any $g \in S_T(\pi)$. Namely, let $S$ be an operator that to any sequence $e_1, \pi_1, g_1, \omega_1, \ldots, e_{T-1}, \pi_{T-1}, g_{T-1}, \omega_{T-1}, e_T$, where $\omega_t \in \Omega$, $\pi_t \in \mathcal{P}(\Omega)$, $g_t \in \mathbb{R}^\Omega$, $t = 1, \ldots, T-1$, and $e_t$, $t = 1, \ldots, T$ are some arbitrary values, assigns a function $S_T : \mathcal{P}(\Omega) \to 2^{\mathbb{R}^\Omega}$ such that $S_T(\pi)$ is a *non-empty* subset of $\mathbb{R}^\Omega$ for all $\pi \in \mathcal{P}(\Omega)$. $S$ is called a *multivalued supermartingale* if for any sequence of arguments where $g_t \in S_t(\pi_t)$, for any $\pi \in \mathcal{P}(\Omega)$, $S_T(\pi) \neq \emptyset$ and for all $g \in S_T(\pi)$ it holds

$$\mathrm{E}_\pi g \leq g_{T-1}(\omega_{T-1}). \qquad (4)$$

A multivalued supermartingale is called *forecast-continuous* if for every $S_T$, the set $\{(\pi, g) \mid \pi \in \mathcal{P}(\Omega), g \in S_T(\pi)\}$ is closed and additionally for every $\pi \in \mathcal{P}(\Omega)$ the set $S_T(\pi) + [0, \infty]^\Omega = \{g \in \mathbb{R}^\Omega \mid \exists g' \in S_T(\pi) \forall \omega \in \Omega \, g'(\omega) \leq g(\omega)\}$ is convex.

Note that if $S$ is a forecast-continuous multivalued supermartingale and $S_t(\pi)$ always consists of exactly one element, $S$ is (equivalent to) a forecast-continuous supermartingale in the former sense: closedness of the graph of $S_T$ means that $S_T(\pi)$ is a continuous function of $\pi$ and the convexity requirement becomes trivial.

### 3.3 Levin's Lemma for Multivalued Supermartingales

Here we prove two version of Levin's lemma suitable for multivalued supermartingales. The first variant (it is simpler) will be used for PEA with finite outcome set $\Omega$. The second variant will be used for DTOL.

**Lemma 2.** *Let $\Omega$ be a finite set. Let $X$ be a compact subset of $\mathbb{R}^\Omega$. Let $q \subseteq \mathcal{P}(\Omega) \times X$ be a relation. Denote $q(\pi) = \{g \mid (\pi, g) \in q\}$ and $\operatorname{ran} q = \cup_{\pi \in \mathcal{P}(\Omega)} q(\pi) \subseteq X$. Suppose that $q$ is closed, for every $\pi \in \mathcal{P}(\Omega)$ the set $q(\pi)$ is non-empty and the set $q(\pi) + [0, \infty]^\Omega$ is convex. If for some real constant $C$ it holds that for every $\pi \in \mathcal{P}(\Omega)$*

$$\forall g \in q(\pi) \quad \mathrm{E}_\pi g \leq C,$$

*then there exists $g \in \operatorname{ran} q$ such that*

$$\forall \omega \in \Omega \quad g(\omega) \leq C.$$

We derive the lemma from Lemma 1 similarly to the derivation of Kakutani's fixed point theorem for multi-valued mappings (see, e.g. [1, Theorem 11.9]) from Brouwer's fixed point theorem. Unfortunately, we did not find a way just to refer to Kakutani's theorem and have to repeat the whole construction with appropriate changes.

*Proof.* Note first that $\mathcal{P}(\Omega)$ is compact for finite $\Omega$, hence $q$ is compact as a closed subset of a compact set. Let $M_q = \max_{g \in \operatorname{ran} q, \omega \in \Omega} |g(\omega)|$.

For every natural $m > 0$, let us take any $(1/m)$-net $\{\pi_k^m\}$ on $\mathcal{P}(\Omega)$ such that for every $\pi \in \mathcal{P}(\Omega)$ there is at least one net element $\pi_k^m$ at the distance less than $1/m$ from $\pi$ and for every $\pi \in \mathcal{P}(\Omega)$ there are at most $4|\Omega|^2$ elements of the net at the distances less than $1/m$ from $\pi$. (One can use here any reasonable distance on $\mathcal{P}(\Omega)$, for example, the maximum absolute value of the coordinates of the difference.) For every $\pi_k^m$ in the net, fix any $g_k^m \in q(\pi_k^m)$ (recall that $q(\pi_k^m)$ is non-empty).

Now let us define a function $q^m \colon \mathcal{P}(\Omega) \times \Omega \to \mathbb{R}$ as a linear interpolation of the points $(\pi_k^m, g_k^m)$. Namely, let $\{u_k^m\}$ be a partition of unity of $\mathcal{P}(\Omega)$ subordinate to $U_{1/m}(\pi_k^m)$, the $(1/m)$-neighborhoods of $\pi_k^m$ (that is, $u_k^m(\pi)$ are non-negative, $u_k^m(\pi) = 0$ if the distance between $\pi$ and $\pi_k^m$ is $1/m$ or more, and the sum over $k$ of all $u_k^m(\pi)$ is 1 at any $\pi$). Let $q^m(\pi, \omega) = \sum_k u_k^m(\pi) g_k^m(\omega)$.

The function $q^m$ is forecast-continuous. Let us find an upper bound on $\mathrm{E}_\pi q^m(\pi, \cdot)$:

$$\mathrm{E}_\pi q^m(\pi, \cdot) = \sum_k u_k^m(\pi) \mathrm{E}_\pi g_k^m$$
$$= \sum_k u_k^m(\pi) \mathrm{E}_{\pi_k^m} g_k^m + \sum_k u_k^m(\pi) \sum_{\omega \in \Omega} (\pi(\omega) - \pi_k^m(\omega)) g_k^m(\omega) \leq C + M_q |\Omega|/m$$

(the bound on the first term holds since $g_k^m \in q(\pi_k^m)$ and hence $\mathrm{E}_{\pi_k^m} g_k^m \leq C$).

By Lemma 1 we can find a point $\pi^m \in \mathcal{P}(\Omega)$ such that

$$\forall \omega \in \Omega \quad q^m(\pi^m, \omega) \leq C + M_q |\Omega|/m \,.$$

Recalling that $q^m(\pi^m, \omega) = \sum_k u_k^m(\pi^m) g_k^m(\omega)$ and that there are at most $4|\Omega|^2$ non-zero values among $u_k^m(\pi^m)$, we get the following statement. There exist some $\alpha_k^m \geq 0$, $k = 1, \ldots, 4|\Omega|^2$, $\sum_k \alpha_k^m = 1$, and some $g_k^m \in q(\pi_k^m)$ with $\pi_k^m$ at the distance at most $1/m$ from $\pi^m$ such that

$$\forall \omega \in \Omega \quad \sum_{k=1}^{4|\Omega|^2} \alpha_k^m g_k^m(\omega) \leq C + M_q |\Omega|/m \,. \tag{5}$$

Since $\mathcal{P}(\Omega)$ is compact, we can find a limit point $\pi^*$ of $\pi^m$. It will be a limit point of $\pi_k^m$ as well. Since $q$ is compact, we can find $g_k^* \in q(\pi^*)$ such that $(\pi^*, g_k^*)$ are limit points of $(\pi_k^m, g_k^m)$ for each $k$. Finally, since $\mathcal{P}(\{1, \ldots, 4|\Omega|^2\})$ is compact, we can find limit points $\alpha_k^*$ (corresponding to the points $g_k^*$).

Taking the limits as $m \to \infty$ over the convergent subsequences in (5), we get

$$\forall \omega \in \Omega \quad \sum_{k=1}^{4|\Omega|^2} \alpha_k^* g_k^*(\omega) \leq C \,.$$

Since $q(\pi^*) + [0, \infty]^\Omega$ is convex, the convex combination $\sum_{k=1}^{4|\Omega|^2} \alpha_k^* g_k^*$ belongs to $q(\pi^*) + [0, \infty]^\Omega$. In other words, the combination is minorized by some $g^* \in q(\pi^*)$ and

$$g^*(\omega) \leq \sum_{k=1}^{4|\Omega|^2} \alpha_k^* g_k^*(\omega) \leq C$$

for all $\omega \in \Omega$. $\square$

Now let us prove a variant of the lemma suitable for the DTOL framework, where the set of outcomes is infinite. Here we make a strong assumption: the supermartingale values $S_T(\pi)$ depend on $\pi$ in a very limited way: just on the mean of $\pi$.

**Lemma 3.** *Let $\Omega$ be $[0,1]^N$. Let $X$ be a compact subset of $\mathbb{R}^\Omega$. Let $q \subseteq \mathcal{P}(\Omega) \times X$ be a relation. Denote $q(\pi) = \{g \mid (\pi, g) \in q\}$ and $\operatorname{ran} q = \cup_{\pi \in \mathcal{P}(\Omega)} q(\pi) \subseteq X$. Assume that if $\int \omega \pi_1(d\omega) = \int \omega \pi_2(d\omega)$ then $q(\pi_1) = q(\pi_2)$. Suppose that $q$ is* *closed, for every $\pi \in \mathcal{P}(\Omega)$ the set $q(\pi)$ is non-empty and the set $q(\pi) + [0, \infty]^\Omega$ is convex. If for some real constant $C$ it holds that for every $\pi \in \mathcal{P}(\Omega)$*

$$\forall g \in q(\pi) \quad \mathrm{E}_\pi g \leq C\,,$$

*then there exists $g \in \operatorname{ran} q$ such that*

$$\forall \omega \in \Omega \quad g(\omega) \leq C\,.$$

*Proof.* Since $[0,1]^N$ is a compact metric space, the space $\mathcal{P}([0,1]^N)$ with weak topology is compact too (see, e.g. [10, Prop. B.28]). Hence $q$ is compact as a closed subset of a compact set. Let $M_q = \max_{g \in \operatorname{ran} q, \omega \in \Omega} |g(\omega)|$.

We consider $\mathcal{P}(\Omega)$ as a metric space with Wasserstein distance $W(\pi, \pi') = \sup_f |\mathrm{E}_\pi f - \mathrm{E}_{\pi'} f|$, where the supremum is taken over 1-Lipschitz functions (see [10, Def. B.20]). For every natural $m > 0$, let us construct a $(1/m)$-net $\{\pi_k^m\}$ on $\mathcal{P}(\Omega)$ with the following property. Let $\omega_i^m$ be a $(1/(2m))$-net on $\Omega$ such that at most $4N^2$ its elements are at the distance less than $1/(2m)$ from any $\omega \in \Omega$. For any $\pi_k^m$ there exists $\omega_i^m$ such that $\int \omega \pi_k^m(d\omega) = \omega_i^m$. (A net with this property exists: note that for any $\pi$ there is a $\pi'$ at the distance at most $1/(2m)$ such that $\int \omega \pi'(d\omega) = \omega_i^m$; it remains to consider a cover of $1/(2m)$-neighborhoods centered in all $\pi$ with the given expected values.) For every $\pi_k^m$ in the net, let us take any $g_k^m \in q(\pi_k^m)$.

Now let us define a function $q^m \colon \mathcal{P}(\Omega) \times \Omega \to \mathbb{R}$ as a linear interpolation of the points $(\pi_k^m, g_k^m)$. Namely, let $\{u_k^m\}$ be a partition of unity of $\mathcal{P}(\Omega)$ subordinate to $U_{1/m}(\pi_k^m)$, the $(1/m)$-neighborhoods of $\pi_k^m$ (that is, $u_k^m(\pi)$ are non-negative, $u_k^m(\pi) = 0$ if the distance between $\pi$ and $\pi_k^m$ is $1/m$ or more, and the sum over $k$ of all $u_k^m(\pi)$ is 1 at any $\pi$). Let $q^m(\pi, \omega) = \sum_k u_k^m(\pi) g_k^m(\omega)$.

The function $q^m$ is forecast-continuous. Let us find an upper bound on $\mathrm{E}_\pi q^m(\pi, \cdot)$:

$$\mathrm{E}_\pi q^m(\pi, \cdot) = \sum_k u_k^m(\pi) \mathrm{E}_\pi g_k^m$$

$$= \sum_k u_k^m(\pi) \mathrm{E}_{\pi_k^m} g_k^m + \sum_k u_k^m(\pi) \left( \int_\Omega g_k^m(\omega) \pi(d\omega) - \int_\Omega g_k^m(\omega) \pi_k^m(d\omega) \right)$$

$$\leq C + M_q/m$$

(the bound on the first term holds since $g_k^m \in q(\pi_k^m)$ and hence $\mathrm{E}_{\pi_k^m} g_k^m \leq C$).

By Lemma 1 we can find a point $\pi^m \in \mathcal{P}(\Omega)$ such that

$$\forall \omega \in \Omega \quad q^m(\pi^m, \omega) \leq C + M_q/m\,.$$

Among $\pi_k^m$ such that $u_k^m(\pi^m)$ is non-zero, there are at most $4N^2$ different expected values. Let us group all $g_k^m$ corresponding to $\pi_k^m$ with a certain expected value. They belong to the same set $q(\pi_k^m)$, thus their convex combination is minorized by another element $\tilde{g}_i^m$ of the same set. Thus we arrive at the following statement: there are some $\alpha_i^m \geq 0$, $i = 1, \ldots, 4N^2$, $\sum_i \alpha_i^m = 1$, and some $\tilde{g}_i^m \in q(\pi_i^m)$ with $\pi_i^m$ at the distance at most $1/m$ from $\pi^m$ such that

$$\forall \omega \in \Omega \quad \sum_{i=1}^{4N^2} \alpha_i^m \tilde{g}_i^m(\omega) \leq C + M_q/m\,. \tag{6}$$

The rest of the proof is the same as in Lemma 2. □

## 3.4 Hoeffding Supermartingale

Here we introduce a specific multivalued supermartingale, or rather a family of supermartingales, that will be used for our main results.

For technical convenience, our definition of supermartingale $S_t$ consists of two parts: a function $G : \mathcal{P}(\Omega) \to 2^\Gamma$, which assigns a set of decisions $G(\pi) \subseteq \Gamma$ to every $\pi \in \mathcal{P}(\Omega)$, and a function $f_t : \Gamma \times \Omega \to \mathbb{R}$. The values of $S_t$ are defined by the formula:

$$S_t(\pi) = \{g \in \mathbb{R}^\Omega \mid \exists \gamma \in G(\pi) \forall \omega \in \Omega \; g(\omega) = f_t(\gamma, \omega)\}. \tag{7}$$

The part $G(\pi)$ depends on the game $(\Omega, \Gamma, \lambda)$ only and does not change from step to step:

$$G(\pi) = \arg\min_{\gamma \in \Gamma} \mathrm{E}_\pi \lambda(\gamma, \cdot) = \{\gamma \in \Gamma \mid \forall \gamma' \in \Gamma \; \mathrm{E}_\pi \lambda(\gamma, \cdot) \leq \mathrm{E}_\pi \lambda(\gamma', \cdot)\}. \tag{8}$$

**Lemma 4.** *Let $(\Omega, \Gamma, \lambda)$ be a game such that its prediction set $\Lambda = \{g \in \mathbb{R}^\Omega \mid \exists \gamma \in \Gamma \forall \omega \in \Omega \; g(\omega) = \lambda(\gamma, \omega)\}$ is a non-empty compact subset of $\mathbb{R}^\Omega$ and $\Lambda + [0, \infty]^\Omega$ is convex. Then the set*

$$G_\Lambda = \{(\pi, g) \in \mathcal{P}(\Omega) \times \Lambda \mid \exists \gamma \in G(\pi) \forall \omega \in \Omega \; g(\omega) = \lambda(\gamma, \omega)\}$$

*is closed and for every $\pi \in \mathcal{P}(\Omega)$ the sets $G(\pi)$ and $G_\Lambda(\pi) = \{g \mid (\pi, g) \in G_\Lambda\}$ are non-empty and the sets $G_\Lambda(\pi) + [0, \infty]^\Omega$ are convex.*

*Proof.* Since $\Lambda$ is non-empty and compact, the minimum of $\mathrm{E}_\pi g$ is attained for every $\pi$, and hence $G(\pi)$ and also $G_\Lambda(\pi)$ is non-empty.

Assume that $g_1, g_2 \in G_\Lambda(\pi) \subseteq \Lambda$ and $\alpha \in [0, 1]$. Then $\alpha g_1 + (1-\alpha) g_2 \geq g \in \Lambda$ since $\Lambda + [0, \infty]^\Omega$ is convex, and $\mathrm{E}_\pi g \leq \mathrm{E}_\pi(\alpha g_1 + (1-\alpha) g_2) = \mathrm{E}_\pi g_1 = \mathrm{E}_\pi g_2$. Hence $g \in G_\Lambda(\pi)$ and thus $G_\Lambda(\pi) + [0, \infty]^\Omega$ is convex.

It remains to show that $G_\Lambda$ is closed. Let $g_i \in G_\Lambda(\pi_i)$ and $(\pi_i, g_i)$ converges to $(\pi, g)$; we need to show that $g \in G_\Lambda(\pi)$. Indeed, $g \in \Lambda$ since $\Lambda$ is compact and $g_i \to g$. Hence $g = \lambda(\gamma, \cdot)$ for some $\gamma \in \Gamma$. To show that $\gamma \in G(\pi)$, let us take any $\gamma' \in \Gamma$ and check that $\mathrm{E}_\pi g \leq \mathrm{E}_\pi g'$, where $g' = \lambda(\gamma', \cdot)$. Clearly, $\mathrm{E}_{\pi_i} g'$ converges to $\mathrm{E}_\pi g'$ since $\pi_i \to \pi$. Also $\mathrm{E}_{\pi_i} g_i$ converges to $\mathrm{E}_\pi g$. Then for any $\epsilon > 0$ we can find sufficiently large $i$ so that $\mathrm{E}_\pi g \leq \mathrm{E}_{\pi_i} g_i + \epsilon$ and $\mathrm{E}_{\pi_i} g' \leq \mathrm{E}_\pi g' + \epsilon$. We have $\mathrm{E}_{\pi_i} g_i \leq \mathrm{E}_{\pi_i} g'$ since $g_i \in L(\pi_i)$. These three inequalities imply that $\mathrm{E}_\pi g \leq \mathrm{E}_\pi g' + 2\epsilon$. Since $\epsilon$ is arbitrary, we have $\mathrm{E}_\pi g \leq \mathrm{E}_\pi g'$. □

Note that for convex bounded compact games the conditions of the lemma are satisfied by definition. For DTOL, the set $\Lambda = \{g \in \mathbb{R}^{[0,1]^N} \mid \exists \vec{p} \in \Delta_N \forall \vec{\omega} \in [0,1]^N \; g(\omega) = \vec{p} \cdot \vec{\omega}\}$ is obviously non-empty and it is compact and convex as a linear image of simplex $\Delta_N$.

Now consider a function $H : \Gamma \times \Omega \to \mathbb{R}$ of the form

$$H(\gamma, \omega) = \mathrm{e}^{\eta(\lambda(\gamma, \omega) - \lambda(\gamma', \omega)) - \eta^2/2}, \tag{9}$$

where parameter $\gamma' \in \Gamma$ and $\eta \geq 0$.

**Lemma 5.** *Let $(\Omega, \Gamma, \lambda)$ be a game, the range of $\lambda$ is included in $[0, 1]$ and $G(\pi)$ is defined by (8). Then for all $\gamma' \in \Gamma$, for all $\eta \leq 0$, for all $\pi \in \mathcal{P}(\Omega)$, and for all $\gamma \in G(\pi)$ it holds*

$$\mathrm{E}_\pi \mathrm{e}^{\eta(\lambda(\gamma, \cdot) - \lambda(\gamma', \cdot)) - \eta^2/2} \leq 1.$$

*Proof.* Since $\lambda(\gamma,\omega) - \lambda(\gamma',\omega) \in [-1,1]$ for any $\gamma$, $\gamma'$ and $\omega$, the Hoeffding inequality (see e.g. [3, Lemma A.1]) implies that

$$\mathrm{E}_\pi \mathrm{e}^{\eta(\lambda(\gamma,\cdot) - \lambda(\gamma',\cdot))} \le \mathrm{e}^{\eta \mathrm{E}_\pi(\lambda(\gamma,\cdot) - \lambda(\gamma',\cdot)) + \eta^2/2}.$$

It remains to note that $\mathrm{E}_\pi(\lambda(\gamma,\cdot) - \lambda(\gamma',\cdot)) \le 0$ by definition of $G(\pi)$. □

Now we can explain what $f_t$ will be used in (7):

$$f_t(\gamma,\omega) = \sum_{k=1}^{K} p_{t,k} H_{t,k}(\gamma,\omega), \qquad (10)$$

where $p_{t,k} \ge 0$ are some weights and $H_{t,k}$ are functions of the form (9) with some parameters $\eta_{t,k}$ and $\gamma_{t,k}$, cf. (11), (15), (18), (21), (22), and (23). The sum may be infinite or it can be even an integral over some measure $p_t(k)$. As in the definition of supermartingale, the index $t$ may hide the dependence on a long sequence of arguments.

**Lemma 6.** *$S_t$ defined by (7), (8), and (10) satisfies the conditions of Lemma 2 if $(\Omega, \Gamma, \lambda)$ is a bounded convex compact game with finite $\Omega$ or the conditions of Lemma 3 if $(\Omega, \Gamma, \lambda)$ is DTOL, where $S_t(\pi)$ is taken for $q(\pi)$ and $\sum_{k=1}^{K} p_{t,k}$ is taken for $C$.*

*Proof.* If $g \in S_t(\pi)$ then $g = f_t(\gamma,\cdot)$ for some $\gamma \in G(\pi)$. Thus we have $\mathrm{E}_\pi g = \sum_{k=1}^{K} p_{t,k} \mathrm{E}_\pi H_{t,k}(\gamma,\cdot) \le \sum_{k=1}^{K} p_{t,k}$ by Lemma 5.

Clearly, $S_t(\pi) \subseteq \Lambda$ and $\Lambda$ is compact, as remarked after Lemma 4. The set $S_t(\pi)$ is non-empty since $G(\pi)$ is non-empty by Lemma 4.

Let $\phi_t(g) = \sum_{k=1}^{K} p_{t,k} \mathrm{E}_\pi \mathrm{e}^{\eta_{t,k}(g - \lambda(\gamma'_{t,k},\cdot)) - \eta_{t,k}^2/2}$. Note that $g \in G_\Lambda(\pi)$ if and only if $\phi(g) \in S_t(\pi)$. Note also that $\phi_t$ is a convex (and hence continuous) function of $g$. Thus, the graph of $S_t$ is closed since $G_\Lambda$ closed and $S_t(\pi) + [0,\infty]^\Omega$ is convex since $G_\Lambda(\pi) + [0,\infty]^\Omega$ is convex.

The condition $S_t(\pi_1) = S_t(\pi_2)$ when $\int \omega \pi_1(d\omega) = \int \omega \pi_2(d\omega)$ for DTOL follows from the equality $\mathrm{E}_\pi \lambda(\vec{\gamma}, \vec{\omega}) = \mathrm{E}_\pi(\vec{\gamma} \cdot \vec{\omega}) = \vec{\gamma} \cdot \mathrm{E}_\pi \vec{\omega}$. □

## 4 Loss Bounds

In this section, we consider applications of the supermartingale technique to obtaining the loss bounds in several different settings. Let us begin with a simple theorem that shows a clean application of the DFA.

**Theorem 7.** *If $T$ is known in advance then the DFA achieves the bound*

$$L_T \le \min_n L_T^n + \sqrt{2T \ln N}$$

*(for DTOL with $N$ actions as well as for PEA with $N$ experts).*

*Proof.* Let $\eta = \sqrt{2(\ln N)/T}$ and

$$f_t(\gamma,\omega) = \sum_{n=1}^{N} \frac{1}{N} \mathrm{e}^{\eta(L_{t-1} - L_{t-1}^n) - \eta^2/2} \times \mathrm{e}^{\eta(\lambda(\gamma,\omega) - \lambda(\gamma_t^n,\omega)) - \eta^2/2}. \qquad (11)$$

At each step $t$, the DFA finds $\gamma_t$ such that $f_t(\gamma_t, \omega) \leq f_{t-1}(\gamma_{t-1}, \omega_{t-1})$ for all $\omega \in \Omega$. Such a $\gamma_t$ exists due to Lemma 6 combined with Lemma 3 for DTOL or Lemma 2 for PEA. Clearly, $f_{t-1}(\gamma_{t-1}, \omega_{t-1}) = \sum_{n=1}^{N} \frac{1}{N} \exp(\eta(L_{t-1} - L_{t-1}^n) - \eta^2/2)$, and we get that the DFA applied to $\{f_t\}$ guarantees that

$$f_t(\gamma_T, \omega_T) = \sum_{n=1}^{N} \frac{1}{N} e^{\eta(L_T - L_T^n) - \eta^2/2} \leq 1 \,.$$

Bounding the sum from below by one additive term, we get the bound. □

This bound is twice as large as the best bound obtained in [3] (see their Theorems 2.2 and 3.7). Our bound is the same as that in Corollary 2.2 in [3].

### 4.1 Bounds on $\epsilon$-Quantile Regret

The bound in Theorem 7 is guaranteed only once, at step $T$ specified in advance. The next bound is uniform, that is, holds for any $T$, and it holds for $\epsilon$-quantile regret for all $\epsilon > 0$.

**Theorem 8.** *For DTOL with $N$ actions, the DFA achieves the bound*

$$\int_0^{1/e} e^{(L_T - L_T^\epsilon)\eta - T\eta^2/2} \frac{d\eta}{\eta \left( \ln \frac{1}{\eta} \right)^2} \leq \frac{1}{\epsilon}, \tag{12}$$

*for any $T$ and any $\epsilon > 0$, where $L_T^\epsilon$ is a value such that at least $\epsilon$-fraction of actions has the loss after step $T$ not greater than $L_T^\epsilon$. In particular, (12) implies for any $\delta \in (0, 1/4)$*

$$L_T \leq L_T^\epsilon + \frac{2}{\sqrt{2 - \delta}} \sqrt{T \ln \frac{1}{\epsilon} + \frac{1}{2} T \ln \frac{1}{\delta} + 2T \ln \ln T} + \max\left\{ 4,400 \ln \frac{1}{\epsilon} \right\}, \tag{13}$$

*which can be further reduced to*

$$R_T^\epsilon \leq \left( 1 + \frac{1}{\ln T} \right) \sqrt{2T \ln \frac{1}{\epsilon} + 5T \ln \ln T} + O\left( \ln \frac{1}{\epsilon} \right). \tag{14}$$

*The bound holds also for PEA; if each of finitely or infinitely many Experts is assigned some positive weight $p_n$, the sum of all weights being not greater than 1, the DFA achieves (12)–(14) with $L_T^\epsilon$ being a value such that the total weight of Experts that have the loss after step $T$ not greater than $L_T^\epsilon$ is at least $\epsilon$.*

*Proof.* We mix all the supermartingales used in (11) over $\eta \in [0, 1/e]$ according to the probability measure

$$\mu(d\eta) = \frac{d\eta}{\eta \left( \ln \frac{1}{\eta} \right)^2}, \quad \eta \in [0, 1/e] \,.$$

We apply the DFA (that is, at each step $t$, find $\gamma_t$ such that $f_t(\gamma_t, \omega) \leq f_{t-1}(\gamma_{t-1}, \omega_{t-1})$ for all $\omega \in \Omega$) to

$$f_t(\gamma, \omega) = \sum_{n=1}^{N} \frac{1}{N} \int_0^{1/e} \frac{d\eta}{\eta \left( \ln \frac{1}{\eta} \right)^2} e^{\eta(L_{t-1} - L_{t-1}^n) - \eta^2/2} \times e^{\eta(\lambda(\gamma, \omega) - \lambda(\gamma_t^n, \omega)) - \eta^2/2} \tag{15}$$

(for PEA with weighed Experts, the term $1/N$ is replaced by $p_n$) and achieve $f_T(\gamma_T, \omega_T) \leq 1$ for all $T$. Bounding the sum from below by the sum of terms where $L_T^n \leq L_T^\epsilon$, we get

$$\int_0^{1/e} e^{\eta(L_T - L_T^\epsilon) - T\eta^2/2} \frac{d\eta}{\eta \left(\ln \frac{1}{\eta}\right)^2} \leq \frac{1}{\epsilon}. \tag{16}$$

Let us estimate the integral. Notice that the exponent $R\eta - T\eta^2/2$ is positive when $0 \leq \eta \leq 2R/T$ and attains its maximum $R^2/(2T)$ at the mid-point of this interval, $\eta = R/T$. Solving the quadratic inequality

$$R\eta - T\eta^2/2 \geq (1/2 - \delta)R^2/T$$

gives

$$\eta \in \left[\frac{R}{T}\left(1 - \sqrt{2\delta}\right), \frac{R}{T}\left(1 + \sqrt{2\delta}\right)\right]$$

($0 < \delta < 1/2$) and so (16) implies

$$e^{(1/2-\delta)R^2/T} \frac{\ln(1 + \sqrt{2\delta}) - \ln(1 - \sqrt{2\delta})}{(\ln(T/R) - \ln(1 + \sqrt{2\delta}))(\ln(T/R) - \ln(1 - \sqrt{2\delta}))} \leq \frac{1}{\epsilon}$$

when $\left(1 + \sqrt{2\delta}\right)R/T \leq 1/e$. If the last condition does not hold and hence $R$ is close to $T$, one can get from (16) that $T < 400 \ln(1/\epsilon)$. Assuming $\delta < 1/4$, we can obtain

$$e^{(2-\delta)R^2/T} \leq \frac{1}{\epsilon\sqrt{2\delta}} \ln^2 \frac{4T}{R}.$$

For $R \geq 4$, we further obtain

$$(2 - \delta)R^2/T \leq \ln \frac{1}{\epsilon} + \frac{1}{2} \ln \frac{1}{\delta} + 2 \ln \ln T,$$

which finally leads to (13). Substituting $\delta = 1/\ln T$, we get (14). □

The bound (14) is not optimal asymptotically in $T$: it grows as $O(\sqrt{T \ln \ln T})$ as $T \to \infty$, instead of $O(\sqrt{T})$. The next theorem gives an asymptotically optimal bound but using a "fake" DFA.

**Theorem 9.** *For DTOL with $N$ actions, there exists a strategy that achieves the bound*

$$L_T \leq L_T^\epsilon + 2\sqrt{T \ln \frac{1}{\epsilon}} + 7\sqrt{T} \tag{17}$$

*for any $T$ and any $\epsilon$, where $L_T^\epsilon$ is a value such that at least $\epsilon$-fraction of actions has the loss after step $T$ not greater than $L_T^\epsilon$.*
*The bound holds also for PEA; if each of finitely or infinitely many Experts is assigned some positive weight $p_n$, the sum of all weights being not greater than 1, the strategy achieves (17) with $L_T^\epsilon$ being a value such that the total weight of Experts that have the loss after step $T$ not greater than $L_T^\epsilon$ is at least $\epsilon$.*

*Proof.* The algorithm in this theorem is not the DFA and does not use supermartingales properly: we use values $f_t(\gamma_t, \omega_t)$ that may increase at some steps and $f_t(\gamma_t, \omega_t) \leq f_{t-1}(\gamma_{t-1}, \omega_{t-1})$ does not hold. Nevertheless, the increases of $f_t$ stay bounded so that it always holds $f_t(\gamma_t, \omega_t) \leq 1$.

Let $1/c = \sum_{i=1}^{\infty} \frac{1}{i^2}$. At step $T$, our algorithm finds $\gamma_T$ such that $f_T(\gamma_T, \omega) \leq C_T$ for all $\omega$, where

$$f_T(\gamma, \omega) = \sum_{n=1}^{N} \frac{1}{N} \sum_{i=1}^{\infty} \frac{c}{i^2} e^{(i/\sqrt{T})(L_{T-1} - L_{T-1}^n) - (i/2\sqrt{T}) \sum_{t=1}^{T-1}(i/\sqrt{t})}$$
$$\times e^{(i/\sqrt{T})(\lambda(\gamma,\omega) - \lambda(\gamma_T^n, \omega)) - (i/\sqrt{T})^2/2} \quad (18)$$

and

$$C_T = \sum_{n=1}^{N} \frac{1}{N} \sum_{i=1}^{\infty} \frac{c}{i^2} e^{(i/\sqrt{T})(L_{T-1} - L_{T-1}^n) - (i/2\sqrt{T}) \sum_{t=1}^{T-1}(i/\sqrt{t})}.$$

For PEA with weighed experts, it is sufficient to replace $1/N$ by $p_n$ in the definitions of $f_T$ and $C_T$.

Note that $f_T$ has the form (10), hence Lemma 6 applies, and due to Lemma 3 or Lemma 2 such a $\gamma_T$ exists.

Let us prove by induction over $T$ that $C_T \leq 1$. It is trivial for $T = 0$, since $L_0 = L_0^n = 0$ and $\sum_{t=1}^{0} = 0$. Assume that $C_T \leq 1$ and prove that $C_{T+1} \leq 1$. By the choice of $\gamma_T$, we know that $f_T(\gamma_T, \omega_T) \leq C_T \leq 1$. Since the function $x^\alpha$ is concave for $0 < \alpha < 1$, we have

$$1 \geq \left(f_T(\gamma_T, \omega_T)\right)^{\sqrt{T}/\sqrt{T+1}}$$
$$= \left( \sum_{n=1}^{N} \frac{1}{N} \sum_{i=1}^{\infty} \frac{c}{i^2} e^{(i/\sqrt{T})(L_T - L_T^n) - (i/2\sqrt{T}) \sum_{t=1}^{T}(i/\sqrt{t})} \right)^{\sqrt{T}/\sqrt{T+1}}$$
$$\geq \sum_{n=1}^{N} \frac{1}{N} \sum_{i=1}^{\infty} \frac{c}{i^2} \left( e^{(i/\sqrt{T})(L_T - L_T^n) - (i/2\sqrt{T}) \sum_{t=1}^{T}(i/\sqrt{t})} \right)^{\sqrt{T}/\sqrt{T+1}} = C_{T+1}.$$

Now it is easy to get the loss bound. Assume that for an $\epsilon$-fraction of Experts their losses $L_T^n$ are smaller than or equal to $L_T^\epsilon$. Then $f_T(\gamma_T, \omega_T)$ can be bounded from below by

$$\epsilon \sum_{i=1}^{\infty} \frac{c}{i^2} e^{(i/\sqrt{T})(L_T - L_T^\epsilon) - (i/2\sqrt{T}) \sum_{t=1}^{T}(i/\sqrt{t})}.$$

Further, bounding the infinite sum by one of the terms, we get

$$e^{(i/\sqrt{T})(L_T - L_T^\epsilon) - (i/2\sqrt{T}) \sum_{t=1}^{T}(i/\sqrt{t})} \leq \frac{1}{\epsilon} \frac{i^2}{c}.$$

Taking the logarithm, using $\sum_{t=1}^{T}(1/\sqrt{t}) \leq 2\sqrt{T}$ and rearranging the terms, we get

$$L_T \leq L_T^\epsilon + \frac{\sqrt{T}}{i} \left( i^2 + \ln \frac{1}{\epsilon} + 2 \ln i + \ln(1/c) \right).$$

Letting $i = \left\lceil \sqrt{\ln(1/\epsilon)} \right\rceil + 1$ and using the estimates $i \leq \sqrt{\ln(1/\epsilon)} + 2$, $1/i \leq 1$, $(\ln i)/i \leq 2$, $(\ln(1/\epsilon))/i \leq \sqrt{\ln(1/\epsilon)}$, and $\ln(1/c) = \ln(\pi^2/6) \leq 1$, we obtain the final bound. □

**Remark 1.** For DTOL and for PEA with the finite number of Experts, the infinite sum over $i$ in the proof can be replaced by the sum up to $\left\lceil \sqrt{\ln N)} \right\rceil + 1$. However, one should keep decreasing weights $c/i^2$: for uniform weights the bound will have an additional term of the form $O((\ln \ln N)/\ln(1/\epsilon))$.

**Remark 2.** Probably, the first bound for $\epsilon$-quantile regret was stated (implicitly) in [9]. More precisely, that paper considered even more general regret notion: Theorem 1 in [9] gives a bound for PEA with weighed experts under the logarithmic loss of the form

$$L_T \leq \sum_{n=1}^{N} u_n L_T^n + \sum_{n=1}^{N} u_n \ln \frac{u_n}{p_n}$$

for any $\vec{u} \in \Delta_N$; $p_1, \ldots, p_N$ are weights of Experts. Here $p_n$ are known to the algorithm in advance, whereas $u_n$ are not known and the bound holds uniformly for all $u_n$. Taking $u_n = 0$ for Experts not from the $\epsilon$-quantile of the best Experts, and uniform $u_n$ over Experts from the $\epsilon$-quantile, we get the bound in terms of $L_T^\epsilon$. It can be easily checked that the strategy in Theorem 9 also achieves the following bound:

$$L_T \leq \sum_{n=1}^{N} u_n L_T^n + 2\sqrt{T \left( \sum_{n=1}^{N} u_n \ln \frac{u_n}{p_n} \right)} + 7\sqrt{T}$$

for any $\vec{u} \in \Delta_N$ and any $T$. In Theorem 8 one can replace $L_T^\epsilon$ by $\sum_{n=1}^{N} u_n L_T^n$ and $\ln(1/\epsilon)$ by $\sum_{n=1}^{N} u_n \ln(u_n/p_n)$ as well.

**Remark 3.** Theorem 9 can be also adapted to discounted regrets of the form $L_T = \sum_{t=1}^{T} (1-\alpha)^{T-t} \lambda(\gamma_t, \omega_t)$ for a known $\alpha$. Then $\epsilon$ in the bound is replaced by $\alpha$, and $L_T^\epsilon$ by $L_T^n = \sum_{t=1}^{T} (1-\alpha)^{T-t} \lambda(\gamma_t^n, \omega_t)$.

### 4.2 Discussion of the Bounds

For a game with $N$ Experts, the best bound, uniform in $T$, is given by [3, Theorem 2.3]:

$$L_T \leq L_T^n + \sqrt{2T \ln N} + \sqrt{\frac{\ln N}{8}}. \qquad (19)$$

The bounds (14) and (17) with $\epsilon = 1/N$ are always worse than (19). In the bound (17) the leading coefficient at $\sqrt{T \ln N}$ is $\sqrt{2}$ times as much. In the bound (14) the coefficient at $\sqrt{T \ln N}$ is the same, but the other terms are larger, and even the asymptotics is worse when $N$ is fixed and $T \to \infty$.

However, it appears that the bound (19) cannot be transferred to $\epsilon$-quantile regret $R_T^\epsilon = L_T - L_T^\epsilon$. The proof of Theorem 2.3 in [3] heavily relies on tracking the loss of only one best Expert, and it is unclear whether the existence of several good (or identical) Experts can be exploited in this proof. The experiments reported in [4] show that algorithms with good best Expert bounds may have rather bad performance when the nominal number of Experts is much greater than the effective number of Experts.

The first (and the only, as far as we know) bound specifically formulated for $\epsilon$-quantile regret is proven for the NormalHedge algorithm in [4, Theorem 1]:

$$L_T \leq L_T^\epsilon + \sqrt{\left(1 + \ln \frac{1}{\epsilon}\right)\left(3(1 + 50\delta)T + \frac{16 \ln^2 N}{\delta}\left(\frac{10.2}{\delta^2} + \ln N\right)\right)}, \quad (20)$$

which holds uniformly for all $\delta \in (0, 1/2)$. Note that this bound depends on the effective number of actions $1/\epsilon$ and at the same time on the nominal number of actions $N$. The latter dependence is weak, but probably prevents the use of NormalHedge with infinitely many Experts.

The main advantage of our bounds in Theorems 8 and 9 is that they are perfectly in terms of the effective number of Experts. In a sense, the DFA does not need to know in advance the number of Experts.

**Remark 4.** To obtain a precise statement about the unknown number of Expert, one can consider the setting where Experts may come at some later steps; the regret to a late Expert is accumulated over the steps after his coming — it is a simple time selection function (see Subsection 4.3), which switches from 0 to 1 only once. Our algorithms and bounds can be easily adapted for this setting: we must consider infinitely many Experts almost all of which are inactive; and then proceed similarly to Theorem 11.

Both our bounds are worse than (20) asymptotically when $\epsilon$ and $N$ are fixed and $T \to \infty$. In this case, the regret term in (20) grows as $\sqrt{3T \ln(1/\epsilon)} + 3T$, whereas in (17) it grows as $\sqrt{4T \ln(1/\epsilon)} + 7\sqrt{T}$ and in (14), the worst bound, it grows as $\sqrt{5T \ln \ln T + 2T \ln(1/\epsilon)}$.

On the other hand, our bounds are better when $T$ is relatively small. The term $\ln \ln T$ is small for any reasonable practical application (e. g., $\ln \ln T < 4$ if $T$ is the age of the universe expressed in microseconds), and then the main term in (14) is $\sqrt{2T \ln(1/\epsilon)}$, which even fits the optimal bound (19). Bound (17) improves over (20) for $T \leq 10^6 \ln^4 N$.

Now let us say a few words about known algorithms for which an $\epsilon$-quantile regret bounds were not formulated explicitly, but can easily be obtained.

The Weighted Average Algorithm, which is used to obtain bound (19), can be analysed in a manner different from [3, Theorem 2.3], see [11]. Then one can obtain the following bound for $\epsilon$-quantile regret:

$$L_T \leq L_T^\epsilon + \frac{1}{c}\sqrt{T} \ln \frac{1}{\epsilon} + c\sqrt{T},$$

where the constant $c > 0$ is arbitrary but must be fixed in advance. If $\epsilon$ is not known and hence $c$ cannot be adapted to $\epsilon$, the leading term is $O(\sqrt{T} \ln \frac{1}{\epsilon})$, which is worse than (17) for small $\epsilon$ (that is, if we consider a large effective number of actions).

For the Aggregating Algorithm [13] (which can be considered as a special case of the DFA for a certain supermartingale, as shown in [5]), the bound can be trivially adapted to $\epsilon$-quantile regret:

$$L_T \leq cL_T^\epsilon + a,$$

where the possible constants $c \geq 1$ and $a$ depend on the loss function. However, in the case of DTOL or arbitrary convex games, the constant $c$ is strictly greater that 1 and the bound may be much worse than (14) and (17) (when $L_T^\epsilon$ grows significantly faster than $\sqrt{T}$). At the same time, this bound is much better when $L_T^\epsilon \approx 0$ (there is at least $\epsilon$ fraction of "perfect" Experts ).

For the standard setting with the known number of Experts, other "small loss" bounds, of the form $L_T \leq L_T^n + O(\sqrt{L_T^n})$, were obtained. The authors of [4] posed an open question whether similar bounds can be obtained if the (effective) number of actions is not known. We left the question open.

## 4.3 Internal Regret and Time Selection Functions

It was shown in [5] and in [7] that the loss bounds obtained by the DFA can be easily transferred to second-guessing experts and sleeping experts models. A second-guessing expert is a (known) function of Learner's decision. Informally, a second-guessing expert explains how Learner could improve (hopefully) his performance. Sleeping experts (or specialists) introduced in [9] may be inactive at some steps, abstaining from announcing their decision (a specialist may decide that the current problem is outside her expertize area). The regret of Learner to a sleeping expert is counted over the steps when the expert was active.

The models similar to second-guessing experts and sleeping experts were studied in DTOL as internal (or wide range) regret and time selection (or activation) functions respectively (see [12] for a review). The internal regret compares Learner's loss not to the loss of a fixed action, but to the loss of a modification rule of the form "Every time Learner selected action $n$ he should have selected $n'$ instead" (more formally, all the weight $\gamma_{t,n}$ assigned to action $n$ should have been appended to $\gamma_{t,n'}$). The wide range regret deals with more general modification rules which may replace each action by some other action. Note that a fixed action $n$ is also a modification rule that suggests to use $n$ instead of any other action.

A time selection function attached to a modification rule assigns a scaling factor from $[0,1]$ to each step. The regret of Learner to this rule is a sum of the regrets at each step weighed by these factors. This weight can be regarded as a degree of specialist's certainty: when the rule is known to be inapplicable for some reason, the weight is zero; and when the rule is partially relevant, the rule agrees for some partial responsibility only.

As has been recently shown [12], an algorithm achieving in DTOL with $N$ action some regret bound with respect to $N$ can be transformed into an algorithm that achieves the same bound with respect to $K$ for $K$ modification rules with attached time selection functions. This gives the best regret bound $O(\sqrt{T \ln K})$.

We show how to extend the results of Theorems 8 and 9 to internal regret and time selection settings. We do not apply the general method of [12], but directly modify our supermartingales and proofs. Remarkably, we need very modest changes.

A modification rule is represented by $N \times N$ *stochastic* matrix $M$: the matrix elements are non-negative and the sum of every column is 1. The (one-step) regret of Learner's decision $\vec{\gamma} \in \Delta_N$ to the modification rule $M$ on the outcome $\vec{\omega} \in [0,1]^N$ is $\vec{\gamma} \cdot \vec{\omega} - (M\vec{\gamma}) \cdot \vec{\omega}$, where $M\vec{\gamma}$ is the product of matrix $M$ and vector-column $\vec{\gamma}$. The total regret after step $T$ on the sequence of outcomes $\vec{\omega}_1, \vec{\omega}_2, \ldots$ of Learner predicting $\vec{\gamma}_1, \vec{\gamma}_2, \ldots$ with respect to a modification rule $M(t)$ with attached time selection function $I(t)$ is

$$R_T = \sum_{t=1}^{T} I(t)\bigl(\vec{\gamma}_t \cdot \vec{\omega}_t - (M(t)\vec{\gamma}_t) \cdot \vec{\omega}_t\bigr)$$

(cf. $R_{H,I,f}$ in [12]).

**Remark 5.** The definition above reflects a slightly more general notion of a modification rule, which allows, for example, the rules that mean "instead of $n$ select at random $n'$ or $n''$ equiprobably". Khot and Ponnuswami [12] do not discuss such rules explicitly, but it appears that their method works for them as well (unless we miss some subtlety in the proof).

First let us obtain an analogue of Theorem 9. We formulate the bound with respect to the effective number of modification rules. It is very probable that the method of [12] also transforms a bound in terms of the effective number of actions into a bound in terms of the effective number of modification rules, but we did not check.

**Theorem 10.** *In DTOL with $N$ actions, let us have $K$ modifications rules $M_k(t)$, each assigning a stochastic $N \times N$ matrix to each step $t$, with attached time selection functions $I_k(t)$ assigning a number from $[0, 1]$. (The modification rule numbered $k$ may arbitrarily change in time and may depend on the whole history, and so is the time selection function.) There is a strategy that achieves the bound*

$$R_T^\epsilon \leq 2\sqrt{T \ln \frac{1}{\epsilon}} + 7\sqrt{T}$$

*for any $T$ and any $\epsilon$, where $R_T^\epsilon$ is a value such that for at least $\epsilon$-fraction of the rules the regret $R_T^k$ of rule $k$ after step $T$ is not less than $R_T^\epsilon$.*

*Proof.* The proof is very similar to the proof of theorem 9. The only change in the algorithm is that in (18) we replace $(\lambda(\vec{\gamma}, \vec{\omega}) - \lambda(\vec{\gamma}_T^n, \vec{\omega})) = \vec{\gamma} \cdot \vec{\omega} - \omega_n$ by $I_k(t)\bigl(\vec{\gamma} \cdot \vec{\omega} - (M_k(t)\vec{\gamma}) \cdot \vec{\omega}\bigr)$ and thus apply the same algorithm with

$$f_T(\gamma, \omega) = \sum_{k=1}^{K} \frac{1}{K} \sum_{i=1}^{\infty} \frac{c}{i^2} e^{(i/\sqrt{T})R_{T-1}^k - (i/2\sqrt{T})\sum_{t=1}^{T-1}(i/\sqrt{t})} \\ \times e^{(i/\sqrt{T})(I_k(T)(\vec{\gamma} \cdot \vec{\omega} - (M_k(T)\vec{\gamma}) \cdot \vec{\omega})) - (i/\sqrt{T})^2/2}. \quad (21)$$

We need to check that the conditions of Lemma 3 are satisfied. It is enough to observe that $I(T) \leq 1$ and that $\exp\left(\left(i/\sqrt{T}\right)\left(I_k(T)(\vec{\gamma} \cdot \vec{\omega} - (M_k(T)\vec{\gamma}) \cdot \vec{\omega})\right)\right)$ is convex in $\vec{\gamma}$, then the proof of the Lemma 6 applies without changes. The loss bound is obtained as in Theorem 9. □

Theorem 8 can be adapted in a similar way. But we formulate another analogue of the theorem: The bound includes the total number of modification rules instead of the the effective number of them, but the regret of each rule $k$ is bounded in terms of the actual activity time (or awake time) $\sum_{t=1}^{T} I_k(t)$ of the rule, not the total time $T$. We do not know whether bounds referring to the awake time were explicitly stated anywhere; however, a bound of this kind can be obtained from bounds that depend on the loss of the rule (or action), as in [2, Theorem 16] or [12, Theorem 5].

**Theorem 11.** *In DTOL with $N$ actions, let us have $K$ modifications rules $M_k(t)$, each assigning a stochastic $N \times N$ matrix to each step $t$, with attached time selection functions $I_k(t)$ assigning a number from $[0,1]$. The DFA achieves the bound*

$$\int_0^{1/e} e^{\eta R_T^k - T_k(T)\eta^2/2} \frac{d\eta}{\eta \left(\ln \frac{1}{\eta}\right)^2} \leq K\,,$$

*where $T_k(T) = \sum_{t=1}^T I_k(t)$, for any $T$ and $k = 1, \ldots, K$. In particular, the above bound implies for any $\delta \in (0, 1/4)$*

$$R_T^k \leq \frac{2}{\sqrt{2-\delta}}\sqrt{T_k(T)\ln\frac{1}{K} + \frac{1}{2}T_k(T)\ln\frac{1}{\delta} + 2T_k(T)\ln\ln T_k(T)}$$
$$+ \max\{4, 400\ln K\}\,,$$

*which can be further reduced to*

$$R_T^\epsilon \leq \left(1 + \frac{1}{\ln T_k(T)}\right)\sqrt{2T_k(T)\ln K + 5T_k(T)\ln\ln T_k(T)} + O(\ln K)\,.$$

*Proof.* We change the supermartingale used for Theorem 8, similarly to the proof of Theorem 10. Namely, we apply the DFA to the supermartingale

$$f_t(\gamma, \omega) = \sum_{k=1}^K \frac{1}{K} \int_0^{1/e} \frac{d\eta}{\eta \left(\ln \frac{1}{\eta}\right)^2} e^{\eta R_{t-1}^k - T_k(t-1)\eta^2/2}$$
$$\times e^{\eta I_k(T)(\vec{\gamma}\cdot\vec{\omega} - (M_k(T)\vec{\gamma})\cdot\vec{\omega}) - (\eta I_k(t))^2/2}\,. \quad (22)$$

Note that in contrast to the proof of Theorem 10, $I_k(t)$ appears also in the "Hoeffding correction term" $e^{-\eta^2/2}$. The rest of the proof does not change much. To get the loss bound we observe that $\sum_{t=1}^T (I_k(t))^2 \leq T_k(T)$ since $I_k(t) \in [0,1]$. □

### 4.4 A Toy Example of a Multiobjective Bound

In this subsection, we discuss bounds with respect to two loss functions. In [7], we showed how to cope with several mixable loss functions. Here we combine a mixable loss function (the square loss) with a non-mixable one (the absolute loss).

Let us describe an informal prediction setting where such a combination of loss functions can make sense. We want to predict the probability of rain. We have two groups of Experts. The first group consists of Metoffices that give the probability and evaluate the result according to the Brier (square) loss function. The second group is Simpletons, they give a boolean ('rain'/'no rain') prediction and count the number of errors (the simple prediction game). We must provide a pair, a probability and a boolean prediction, and the two components of our prediction must agree in the following sense: if we give probability of rain more than one half, we must predict 'rain'; if we give probability of rain less than one half, we must predict 'no rain'; only if we give the probability $1/2$, we may choose the boolean prediction arbitrary (so we can randomize here). In the theorem below we bound both Learner's Brier loss and Learner's expected (over the internal randomizer) number of errors.

**Theorem 12.** *Assume that we are given $K$ Experts that give predictions $p^k \in [0,1]$ and $M$ Experts that give predictions $b^m \in \{0,1\}$. Learner is allowed to give predictions $(p, \tilde{p}) \in [0,1] \times [0,1]$, with the following restriction: if $p < 1/2$ then $\tilde{p} = 0$ and if $p > 1/2$ then $\tilde{p} = 1$. Then there exists a strategy for Learner guaranteeing for any sequence of outcomes $\omega_1, \omega_2, \ldots$ that for any $T$ and for any $k$ it holds*

$$\sum_{t=1}^{T}(p_t - \omega_t)^2 \leq \sum_{t=1}^{T}(p_t^k - \omega_t)^2 + \frac{1}{2}\ln(K+M),$$

*and for any $T$ and for any $m$ it holds*

$$\sum_{t=1}^{T}|\tilde{p}_t - \omega_t| \leq \sum_{t=1}^{T}[b_t^m \neq \omega_t] + O(\sqrt{T\ln(K+M)} + T\ln\ln T),$$

*where $[b_t^m \neq \omega_t] = 1$ if $b_t^m \neq \omega_t$ and $[b_t^m \neq \omega_t] = 0$ otherwise.*

*Proof.* Let $A = \{(p, \tilde{p}) \in [0,1]^2 \mid \tilde{p} = 0 \text{ if } p < 1/2 \text{ and } \tilde{p} = 1 \text{ if } p > 1/2 \text{ and }\} = \{(p, 0) \mid p \in [0, 1/2)\} \cup \{(1/2, \tilde{p}) \mid \tilde{p} \in [0,1]\} \cup \{(p, 1) \mid p \in (1/2, 1]\}$. We apply the DFA to supermatingale $S_T$ on $\Omega = \{0,1\}$ defined by (7) with

$$f_T(p, \tilde{p}, \omega) = \frac{1}{K+M}\sum_{k=1}^{K} e^{2\sum_{t=1}^{T}((p_t - \omega_t)^2 - (p_t^k - \omega_t)^2)} \times e^{2((p-\omega)^2 - (p_T^k - \omega)^2)}$$

$$+ \frac{1}{K+M}\sum_{m=1}^{M}\int_{0}^{1/e}\frac{d\eta}{\eta\left(\ln\frac{1}{\eta}\right)^2} e^{\eta\sum_{t=1}^{T}(|\tilde{p}_t - \omega_t| - [b_t^m \neq \omega_t]) - \eta^2/2} \times e^{\eta(|\tilde{p} - \omega| - [b_T^m \neq \omega]) - \eta^2/2} \tag{23}$$

and $G(\pi) = \{(p, \tilde{p}) \in A \mid p = \pi(1)\}$. To ensure that $S_T$ is a supermartingale, we need to check that $\mathrm{E}_\pi\left(|\tilde{p} - \omega| - [b_T^m \neq \omega]\right) \leq 0$ if $(\pi(1), \tilde{p}) \in G(\pi)$. Then we can refer to Lemma 6 and [5, Lemma 2].

Indeed, $\mathrm{E}_\pi\left(|\tilde{p} - \omega| - [b_T^m \neq \omega]\right) = \pi(1)(1 - \tilde{p} - (1 - b_T^m)) + \pi(0)(\tilde{p} - b_T^m) = (\pi(0) - \pi(1))(\tilde{p} - b_T^m)$. If $\pi(1) > 1/2$ then $\pi(0) < 1/2$ and $\tilde{p} = 1 \geq b_T^m$. If $\pi(1) < 1/2$ then $\pi(0) > 1/2$ and $\tilde{p} = 0 \leq b_T^m$. If $\pi(1) = 1/2$ then $\pi(0) = 1/2$. Obviously, in all the cases $(\pi(0) - \pi(1))(\tilde{p} - b_T^m) \leq 0$.

The bound follows in the usual way (cf. Theorem 8). □

**Remark 6.** Let us discuss how to find the numbers $p$ and $\tilde{p}$ such that $f_T(p, \tilde{p}, 0) \leq 1$ and $f_T(p, \tilde{p}, 1) \leq 1$. Consider $x \in [0, 2]$ and two functions

$$p(x) = \begin{cases} x, & \text{if } x < 1/2, \\ 1/2, & \text{if } x \in [1/2, 3/2], \\ x - 1, & \text{if } x > 3/2, \end{cases}$$

and

$$\tilde{p}(x) = \min\{1, \max\{x - 1/2, 0\}\}.$$

Clearly, $p(x)$ and $\tilde{p}(x)$ are continuous functions of $x$. Let

$$g(x, \omega) = f_T(p(x), \tilde{p}(x), \omega) - 1.$$

It is obvious that if $g(x_0, 0) \leq 0$ and $g(x_0, 1) \leq 0$ then we can take $p(x_0)$ and $\tilde{p}(x_0)$ as $p$ and $\tilde{p}$ we are looking for. The supermartingale property of $S_T$ and the definition of $S_T$ imply that

$$p(x)g(x, 1) + (1 - p(x))g(x, 0) \leq 0\,.$$

Substituting $x = 0$, we get $g(0, 0) \leq 0$. Substituting $x = 2$, we get $g(2, 1) \leq 0$. If $g(0, 1) \leq 0$ or $g(2, 0) \leq 0$, we can take $x_0 = 0$ or $x_0 = 2$ respectively. Otherwise, consider the function $\phi(x) = g(x, 1) - g(x, 0)$. It is continuous, $\phi(0) > 0$ and $\phi(2) < 0$, hence there exists $x_0$ such that $\phi(x_0) = 0$. Clearly, $g(x_0, 0) = g(x_0, 1) \leq 0$.

## Acknowledgements

This work was supported by EPSRC, grant EP/F002998/1. We are grateful to Yura Kalnishkan for discussions.